# Machine Learning for Modeling Wireless Radio Metrics with Crowdsourced Data and Local Environment Features

Yifeng Qiu[(1)] and Alexis Bose[(1) †]
Communications Research Centre, Ottawa, Canada, {evanyifeng.qiu, alexis.bose}@ised-isde.gc.ca

*Abstract*—This paper presents a suite of machine learning models, CRC-ML-Radio Metrics, designed for modeling RSRP, RSRQ, and RSSI wireless radio metrics in 4G environments. These models utilize crowdsourced data with local environmental features to enhance prediction accuracy across both indoor at elevation and outdoor urban settings. They achieve RMSE performance of 9.76 – 11.69 dB for RSRP, 2.90 – 3.23 dB for RSRQ, and 9.50 – 10.36 dB for RSSI, evaluated on over 300,000 data points in the Toronto, Montreal, and Vancouver areas. These results demonstrate the robustness and adaptability of the models, supporting precise network planning and quality of service optimization in complex Canadian urban environments.

## I. INTRODUCTION

Radio metrics modeling is central to optimizing quality of service (QoS) in wireless communication systems, particularly in urban environments where signal conditions are complex. To achieve accurate radio metrics predictions, it is essential to leverage radio frequency (RF) propagation models, which provide foundational insights into signal behavior across different environments, enabling radio metrics modeling to deliver desirable predictions that support effective network planning and leading to improvements for mobile users. Existing propagation modeling, such as ITU-R P.2109 [1] and COST Action CA15104 IRACON [2], has addressed various aspects of outdoor-to-indoor signal propagation. ITU-R P.2109 provides standardized methods for predicting building entry loss by estimating signal attenuation at entry points, and it does not extensively cover propagation within buildings or elevation effects. COST Action CA15104 IRACON promotes comprehensive modeling across diverse environments, including indoor and outdoor settings, but remains in an evolving state as standards need to be developed further.

In recent years, the empirical propagation model based on 3GPP standard, 3GPP TR 38.901 [8], has gained traction for propagation in urban macro (UMa) and urban micro (UMi) scenarios. This propagation model considers outdoor-to-indoor signal penetration loss and provides flexibility in accounting for height variations. While TR 38.901 offers detailed modeling for UMa and UMi environments, it still falls short in fully capturing the variability and elevation differences that occur within complex indoor settings. An initial investigation into 4G reference signal received power (RSRP) prediction and generalization has been presented in [19].

In this paper, we propose multiple machine learning (ML) models, namely CRC-ML-4G radio metric models, that apply regression techniques to predict wireless radio metrics in 4G environments, specifically targeting RSRP, reference signal received quality (RSRQ) and received signal strength indicator (RSSI). These three metrics are commonly used to assess the quality and reliability of 4G wireless networks and are critical for QoS optimization. These models leverage crowdsourced user equipment (UE) measurement data spanning multiple frequency bands, Canadian cities and many towers, incorporating local environmental features to enhance 4G radio metrics modeling in diverse urban contexts. By integrating crowdsourced data and engineered features, this approach enables accurate predictions of RSRP, RSRQ, and RSSI in both indoor and outdoor 4G settings. The CRC-ML-4G radio metric models further extends the research work introduced by the Technical University of Denmark [3] – [6], incorporating dynamic, real-world data from crowdsourced sources rather than relying solely on static propagation models or controlled drive tests. This approach allows for comprehensive modeling across varied geographic settings, improving predictive performance for these radio metrics in urban environments.

## II. METHODOLOGY

The CRC-ML-4G radio metric models are designed in two stages: a path loss model is used to generate initial RSRP estimates, followed by a target correction through a deep neural network (DNN). The initial RSRP estimates are intended to accelerate the model's learning process by providing a superior starting point for hyperparameter optimization.

*A. System Pipeline and Model Architecture*

The system pipeline and model architecture for modeling RSRP (CRC-ML-4G-RSRP) is illustrated in Fig. 1. Initially, a path loss estimate is generated by a path loss model based on the input features listed in Table I, including the downlink frequency, distance to transmitter, transmitter height, UE altitude, line-of-sight (LOS) / non-line-of-sight (NLOS) classification, transmit signal penetration into the building and cumulative obstacle distance.

A CRC machine learning-based path loss model (CRC-MLPL 3-feature) [10] combined with the free-space path loss (FSPL) model is used as the initial path loss estimate in this proposed CRC-ML-4G-RSRP modeling. The path loss estimate, along with transmitter power and the number of 4G resource blocks $N$, is used to produce the RSRP estimate, which is further refined by a correction neural network. The correction neural network is trained on a broader set of engineered features listed in Table II. These engineered features are employed to improve each model's ability to predict radio metrics across diverse conditions, including both indoor and outdoor environments, as well as LOS and NLOS scenarios. Correction from the neural network combines with the RSRP estimate to finally derive the RSRP prediction as the output of CRC-ML-4G-RSRP model. This proposed system pipeline and model architecture are also applied to the modeling of RSRQ (CRC-ML-4G-RSRQ) and RSSI (CRC-ML-4G-RSSI) addressed in this paper.

---

[†] These authors contributed equally to this work.

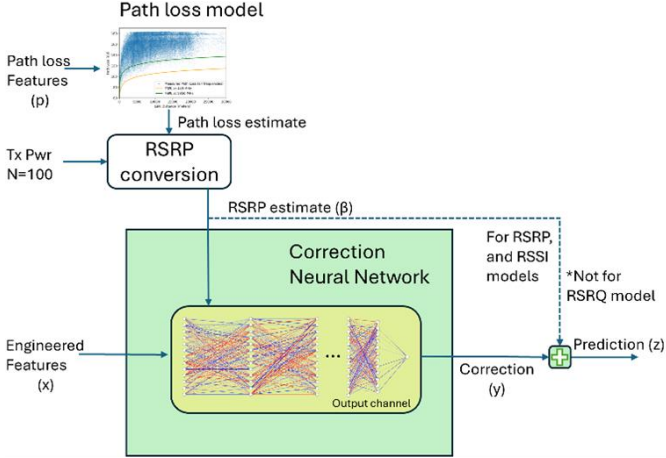

Figure 1. The system pipeline and model architecture for modeling RSRP radio metric. Feature engineering preprocessing is applied to the input crowdsourced data before they are passed to the correction neural network along with an RSRP estimate. The correction output, together with the RSRP estimate, results in the final RSRP prediction. The same pipeline and model architecture are used for modeling the RSSI radio metric. For RSRQ modeling, the RSRP estimate is only used for driving the correction neural network, and is not used to create the final RSRQ prediction.

TABLE I. FEATURES TO THE PATH LOSS MODEL FOR ESTIMATE.

| Path Loss Features | Description |
| --- | --- |
| downlink_frequency | Mobile connection downlink frequency of UE (700 MHz to 2.68 GHz). |
| distance_to_transmitter_km | Straight line distance between transmitter and UE (0 – 3 km). |
| tx_height_m | Transmitter / tower height (5 to 120 m). |
| ue_altitude_ag_m | Altitude of UE (0 to 325 m). |
| line_of_sight | Geometric LOS / NLOS binary classification. |
| building_penetration_length_m | Penetration length of transmit signal into the building (m). |
| total_obstruction_length_3d_m | Cumulative distance of all obstacles the transmitted signal path crosses (m). |

The CRC-ML-4G-RSRP model pipeline and architecture can be formalized as follows:

$$\beta(p) = TxPwr - \alpha(p) - 10 \cdot log_{10}^{12 \cdot N} \quad (1)$$

$$Z(x, w, \theta) = y[x, \beta(p), w, \theta] + \beta(p) \quad (2)$$

where $\beta(p)$ is the RSRP estimate derived from the transmitter power $TxPwr$, number of resource blocks $N$, and path loss estimate $\alpha(p)$ by the path loss model. $Z(x, w, \theta)$ is the RSRP prediction derived from the output of correction neural network $y[x, \beta(p), w, \theta]$ and the RSRP estimate $\beta(p)$. $N$ is set to 100 for radio metrics modeling in the paper. $p$ represents the path loss features, $x$ the engineered features, $w$ the adaptive weights in correction neural network, and $\theta$ the hyperparameters for the model training. The path loss and engineered features are standardized prior to the model training.

TABLE II. ENGINEERED FEATURES TO THE CORRECTION NEURAL NETWORK.

| Engineered Features (RSRP, RSRQ, RSSI) | Description |
| --- | --- |
| downlink_frequency | Mobile connection downlink frequency of UE (700 MHz to 2.68 GHz). |
| distance_to_transmitter_km | Straight line distance between transmitter and UE (0 – 3 km). |
| distance_x_km | Longitude distance between transmitter and UE (km). |
| distance_y_km | Latitude distance between transmitter and UE (km). |
| tx_height_m | Transmitter / tower height (5 to 120 m). |
| tx_power | Transmitter power (29 to 70 dBm). |
| ue_altitude_ag_m | Altitude of UE (0 to 325 m). |
| line_of_sight | Geometric LOS / NLOS binary classification. |
| building_penetration_length_m | Penetration length of transmitted signal into the building (m). |
| building_intersection_count_3d | Building count in the signal path between transmitter and UE. |
| total_obstruction_length_3d_m | Cumulative distance of all obstacles the transmitted signal path crosses (m). |
| **Additional RSRQ and RSSI Features** | |
| day_of_week | Indicating the day in a week (0 = Monday, 1 = Tuesday, ..., 6 = Sunday), only used for modeling RSRQ. |
| hour_of_day | Indicating the hour in a day (0 – 23), only used for modeling RSRQ. |

### B. Data Preparation

The training dataset consists of a combination of open and closed-source data, which has undergone extensive cleaning and integration. To maintain measurement consistency, the UE with the most measurements has been selected.

The closed-source data includes UE crowdsourced radio metrics from the year of 2020 to 2022 and 4G transmitter locations. The open-source data consists of the 4G transmission license database [15], clutter data from OpenStreetMap contributors [16], High Resolution Digital Elevation Model (HRDEM) [17], and data from the City of Toronto [18].

As the crowdsourced data lacks direct RSSI measurements, the RSSI target is derived from the RSRP and RSRQ measurements according to the definition outlined in 3GPP TR 36.214 [20].

Before modeling, the spatial components of data were visually verified after being converted into .kml format, as shown in Fig. 2. The training dataset was created without the information regarding whether the UE was moving or stationary, nor about its data usage during the measurements. The absence of those two features rendered the data non-ideal, as highlighted by [7] and [9], respectively. Consequently, only temporal and spatial data were utilized at the time of the radio metric measurements. This approach enables the CRC-ML-4G radio metric suite to model general trends of the real world, making it practical for network optimization, planning, and spectrum regulation.

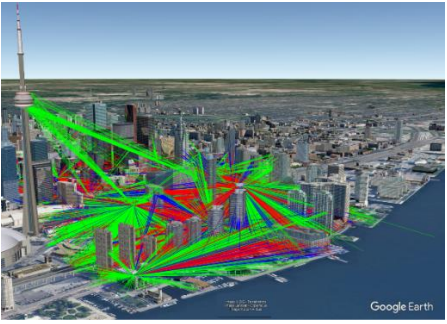

Figure 2. Toronto KML line-of-sight (LOS) / non-line-of-sight (NLOS) results for visual verification. Red is NLOS and green LOS, whereas blue is LOS to be further verified.

*C. Crowdsourced Dataset for Training and Testing*

Each CRC-ML-4G radio metric model (RSRP, RSRQ, RSSI) is trained and blindly evaluated on the combined dataset of both indoor and outdoor data, without separating these environments, ensuring the capability to generalize effectively across diverse environment settings. The crowdsourced datasets used for training, validation and blind testing by the CRC-ML-4G radio metric models are listed in Table III.

Each model is trained and validated with a Greater Toronto Area (GTA) dataset, excluding the data of dpz833 geohash6 [11] area, where geohash6 refers to a 6-character alphanumeric string representing a geographic area with a precision of approximately 0.61 km × 0.61 km. The model is then blindly tested with the dpz833 dataset. Generalization is verified through blind tests on the Montreal and Vancouver datasets.

TABLE III. CROWDSOURCED DATASETS FOR CRC-ML-4G RADIO METRICS MODELING.

| Dataset | Counts Total | Outdoor | Indoor | Description |
|---|---|---|---|---|
| Train | 210,944 | 128,464 | 82,480 | GTA (without dpz833) |
| validation | 101,963 | 43,772 | 58,191 | GTA (without dpz833) |
| blind test 1 | 1,293 | 1,050 | 243 | dpz833 |
| blind test 2 | 165,613 | 110,001 | 55,612 | Montreal |
| blind test 3 | 198,075 | 122,463 | 75,612 | Vancouver |

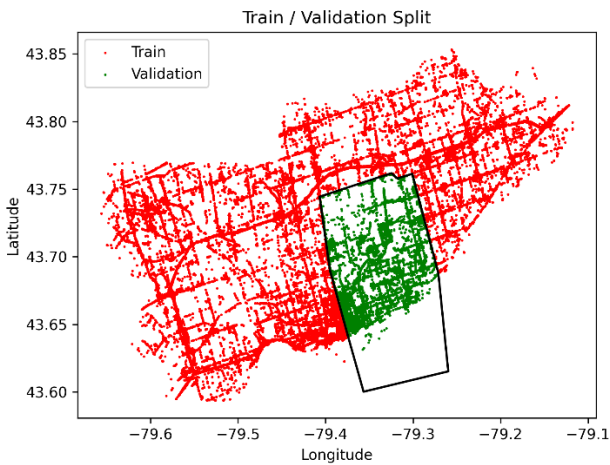

Figure 3. Crowdsourced dataset for the CRC-ML-4G radio metric models train and validation in Greater Toronto Area (GTA) excluding dpz833. The dataset, sized at 312,907 data samples, is geographically split along Yonge Street, York Mills Road, and Warden Avenue as boundaries, with 67.4% of the data used for train and 32.6% for validation.

Using geographically distinct datasets for the train and validation based on different geohash6 areas is to prevent data leakage and evaluate the adaptability and robustness of the model across different geographic regions. Fig. 3 illustrates a train and validation split of crowdsourced dataset used by the CRC-ML-4G radio metric models.

## III. RESULTS

The CRC-ML-4G model architecture is implemented with the PyTorch framework. A unified framework for scalable and distributed computing, Ray [12], is leveraged to speed up the model training, assist in the model architecture search and tuning, and optimize the hyperparameters using an optimizer HyperOptSearch [13] for batch size, neural network layers, weight decay, epochs and learning rate. In this paper, the trained model is determined based on an extensive search involving 300 experiments using a cluster of twenty Amazon EC2 g5.4xlarge instances that run the training process for over 14 hours. A mean squared log-scaled error (msle) loss [14] is implemented in the neural network with an empirically determined alpha value of 5, in order to prevent the model from frequently predicting the mean value.

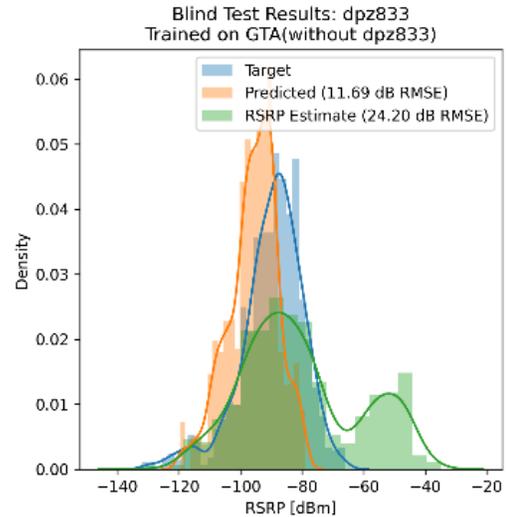

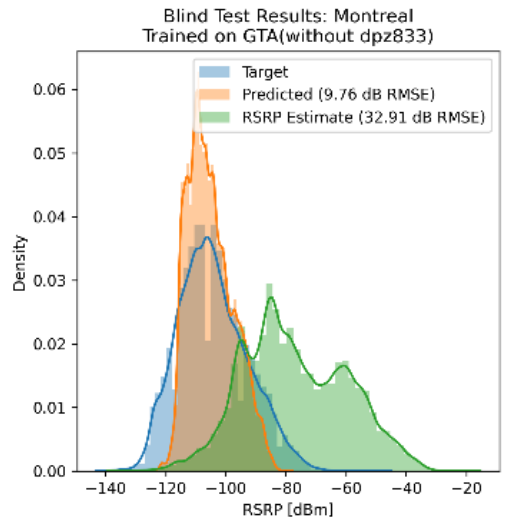

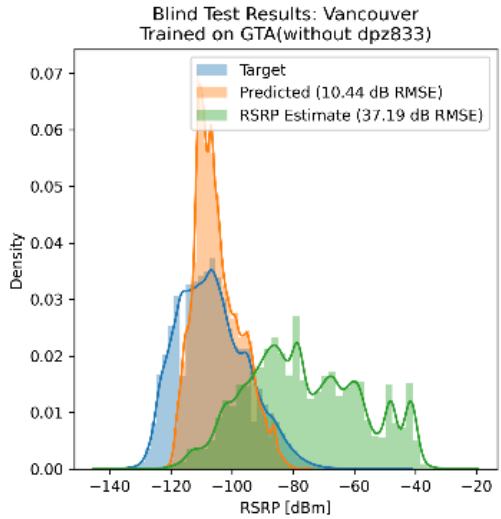

Figure 4. Blind test results of CRC-ML-4G-RSRP for dpz833, Montreal and Vancouver, containing both indoor and outdoor settings. The kernel density estimates depict the RSRP distributions of the measured data (blue), the predicted result from the trained model (orange) and the initial RSRP estimate (green).

The blind test results for the CRC-ML-4G-RSRP model are shown in Fig. 4. These results are generated based on the RMSE, with the best-trained model applied to dpz833, Montreal, and Vancouver. In this figure, the blue curve represents the kernel density of the target measurement data. The green curve shows the kernel density of the initial RSRP estimate generated by the CRC-MLPL 3-feature model [10] combined with the FSPL model, prior to refinement by the correction neural network. The orange curve illustrates the predicted RSRP from the CRC-ML-4G-RSRP model. The RMSE performances of both the initial RSRP estimate and the predicted RSRP are evaluated against the target measurements. These blind test RMSE results for the CRC-ML-4G-RSRP model reflect the consistency and reliability of the trained model in different scenarios, signifying the model strength in handling complex signal propagations that involve both LOS and NLOS conditions in indoor and outdoor settings, contributing to more accurate RSRP predictions.

The blind test results for CRC-ML-4G-RSRQ and CRC-ML-4G-RSSI models across the dpz833, Montreal and Vancouver are shown in Fig. 5 and Fig. 6.

In Fig. 5, the kernel density of the target RSRQ measurements exhibits a segmented pattern due to the discrete nature of RSRQ values in the crowdsourced data (e.g., -3, -4, -5, …). The segmented pattern reflects the inherent quantization of RSRQ measurements in the dataset, rather than a continuous distribution. The blind test results of RSRQ modeling reveal a high degree of alignment, in terms of RMSE performance, between the predicted values and the measured data across all test areas. The inclusion of two temporal features to the CRC-ML-4G-RSRQ model, i.e., *day_of_week* and *hour_of_day*, has been instrumental in enhancing RSRQ prediction accuracy. These features allow the model to capture the natural fluctuations in signal quality that occur at different times of the day and across days of the week. This temporal sensitivity enables the model to account for daily and hourly variations in the spectrum environment which impact RSRQ metric, leading to the more accurate and context-aware predictions.

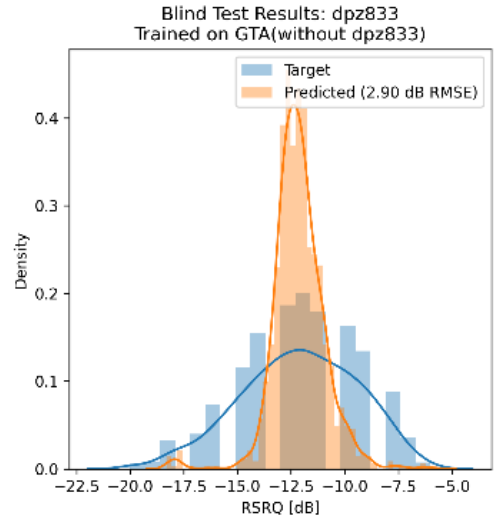

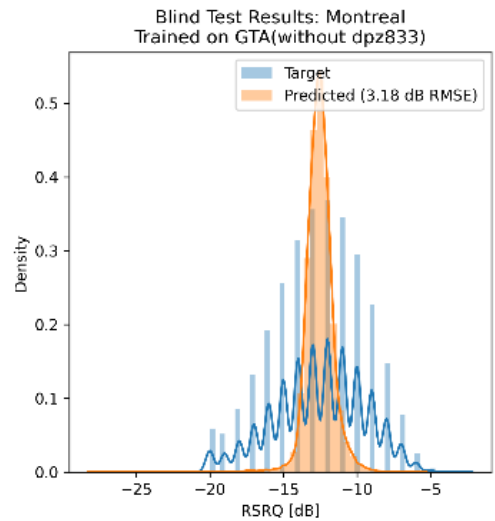

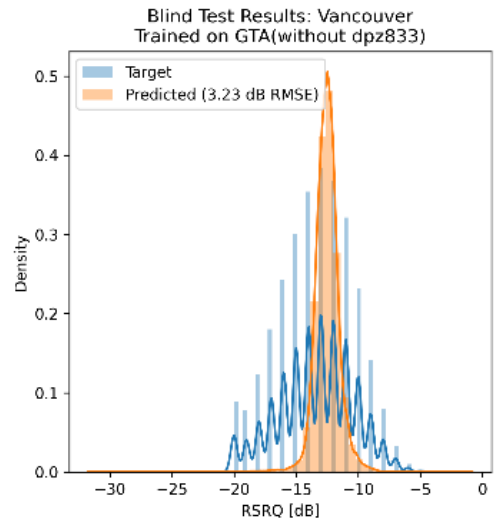

Figure 5. Blind test results of CRC-ML-4G-RSRQ for dpz833, Montreal and Vancouver, containing both indoor and outdoor settings. The kernel density estimates illustrate the RSRQ distributions of the measured data (blue) and the predicted result from the trained model (orange). In this case, the initial RSRP estimate is not directly used to create the final RSRQ prediction.

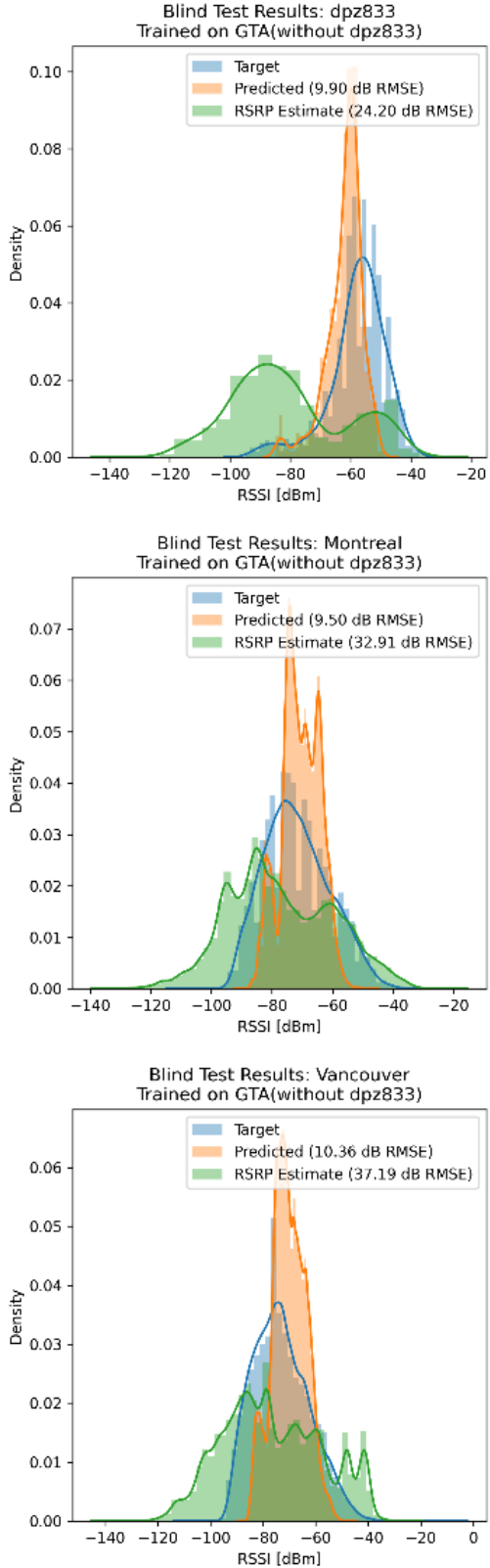

Figure 6. Blind test results of CRC-ML-4G-RSSI for dpz833, Montreal and Vancouver, containing both indoor and outdoor settings. The kernel density estimates present the RSSI distributions of the measured data (blue), the predicted result from the trained model (orange) and the initial RSRP estimate (green).

Fig. 6 presents the CRC-ML-4G-RSSI model blind test results, where the model aligns well with the measured RSSI distributions across all three (dpz833, Montreal, Vancouver) geographic datasets. This model is also sensitive to temporal changes, ensuring consistent RSSI estimations that adapt to varying environmental conditions. This consistency across diverse datasets further validates the robustness of the CRC-ML-4G radio metric models in accurately predicting wireless radio metrics under various urban scenarios.

Table IV generates a summary of the RMSE performance achieved by the CRC-ML-4G radio metric models in predicting RSRP, RSRQ, and RSSI across different urban areas.

For RSRP predictions, RMSE values range from 9.76 to 11.69 dB overall, with higher RMSE observed for indoor samples in dpz833 due to a limited sample size of only 295 indoor data points for test. RSRQ predictions achieve consistent RMSE values around 3 dB across all test areas. RSSI predictions yield RMSE values between 9.50 and 10.36 dB overall, showing the model's adaptability for both indoor and outdoor environments. The RSSI and RSRQ models, having been trained with temporal features, exhibit the capacity to effectively capture and model temporal noises.

Recall that the CRC-ML-4G models are seeded by the MLPL model, which has an RMSE of 7.46 dB when predicting path loss using UK urban datasets, presented in the 3-feature MLPL Fully-Connected Network (FCN) [Table III, 10]. Path loss is the primary factor causing signal weakening in radio communications, which effectively establishes the performance limit for radio signal prediction models. As a result, the CRC-ML-4G models cannot perform better than this fundamental constraint. Therefore, developing models that come within 2 – 4 decibels of the path loss error is considered a modeling achievement.

Consequently, the CRC-ML-4G radio metric models are able to achieve the performance only 2 – 4 dB worse than the path loss error, which is desirable for the accurate modeling. In practice, this minor deviation indicates that the CRC-ML-4G radio metric models maintain high accuracy on modeling different radio metrics when applied to the diverse Canadian urban datasets.

TABLE IV. SUMMARY OF THE MODELING PERFORMANCES.

| City / geohash6 for Blind Test | | CRC-ML-4G-RSRP RMSE(dB) | CRC-ML-4G-RSRQ RMSE(dB) | CRC-ML-4G-RSSI RMSE(dB) |
|---|---|---|---|---|
| dpz833 | Indoor | 14.68 | 3.15 | 12.22 |
| | Outdoor | 10.68 | 2.82 | 9.12 |
| | Overall | 11.69 | 2.90 | 9.90 |
| Montreal | Indoor | 8.87 | 2.91 | 9.01 |
| | Outdoor | 10.18 | 3.30 | 9.74 |
| | Overall | 9.76 | 3.18 | 9.50 |
| Vancouver | Indoor | 9.85 | 2.95 | 10.67 |
| | Outdoor | 10.79 | 3.39 | 10.16 |
| | Overall | 10.44 | 3.23 | 10.36 |

Table V provides an overview of the CRC-ML-4G model architectures used for modeling RSRP, RSRQ, and RSSI metrics, detailing the neural network configurations specific to each radio metric. These model architectures are tailored to optimize performance, with varying numbers of hidden layers, neurons and output channels, reflecting the different complexity requirements of each radio link metric to enhance the ML model predictive performance and generalizability across multiple datasets for different geographic areas.

TABLE V. THE CRC-ML-4G MODEL ARCHITECTURES FOR MODELING RSRP, RSRQ AND RSSI RADIO METRICS. FOR EACH MODEL ARCHITECTURE, THE NUMBER OF HIDDEN NEURONS, LAYERS AND OUTPUT CHANNELS ARE GIVEN.

| Radio Metrics | Neural Network Layer | Output Channel | Batch Size | Learning Rate | Weight Decay |
|---|---|---|---|---|---|
| RSRP | [64, 64, 64, 64, 64, 64] | [64, 256, 10, 1] | 69 | 3.03e-2 | 2.07e-4 |
| RSRQ | [64, 64, 64, 64, 64, 64, 64, 64] | [64, 32, 10, 1] | 55 | 6.06e-6 | 3.8e-3 |
| RSSI | [16, 16, 16, 16, 16] | [16, 512, 10, 1] | 52 | 6.29e-4 | 3.17e-7 |

## IV. CONCLUSION

This paper presents a robust suite of machine learning radio metric models, CRC-ML-4G, which is capable of modeling wireless radio metrics for RSRP, RSRQ and RSSI in both indoor and outdoor environments. The CRC-ML-4G models refine an initial RSRP estimate derived from a path loss model, achieving an RMSE performance of 9 – 11dB. This performance is only 2 – 4 dB above the performance limit defined by the path loss error. Additionally, the blind test results across multiple urban datasets demonstrate the generalizability and suitability of CRC-ML-4G radio metric models for diverse indoor and outdoor urban settings.

Future work will focus on incorporating local imagery to further refine environmental feature inputs and evaluate the model with additional cities and technologies. These enhancements aim to extend the adaptability and scalability of the models, reinforcing their potential as valuable tools for optimizing wireless network planning and performance across varied geographic and environmental conditions.

ACKNOWLEDGMENT

The authors would like to thank the CRC colleagues for their contributions to this work, particularly Scott Campbell and Siva Palaninathan for their expert insights and great support throughout the project. We also acknowledge Tutela as the source of the crowdsourced data and OpenStreetMap contributors for supplying open-source mapping data used in this study. We express our gratitude for their collaborative efforts in maintaining such a vital resource.